\documentclass{article} 
\usepackage{iclr2017_conference,times}
\usepackage{algorithm2e}
\usepackage{graphicx}
\usepackage{url}

\title{DyVEDeep: Dynamic Variable Effort \\ Deep Neural Networks}

\author{Sanjay Ganapathy \\
Department of Computer Science and Engineering\\
Indian Institute of Technology Madras\\
Chennai, Tamil Nadu, India \\
\texttt{sanjaygana@gmail.com} \\
\And
Swagath Venkataramani \thanks{Currently a Research Staff Member at IBM T.J. Watson Reseach Center, Yorktown Heights, NY} \\
Department of Electrical and Computer Engineering \\
Purdue University \\
West Lafayette, IN, USA \\
\texttt{venkata0@purdue.edu} \\
\And
Balaraman Ravindran \\
Department of Computer Science and Engineering\\
Indian Institute of Technology Madras\\
Chennai, Tamil Nadu, India \\
\texttt{ravi@cse.iitm.ac.in} \\
\And
Anand Raghunathan \\
Department of Electrical and Computer Engineering \\
Purdue University \\
West Lafayette, IN, USA \\
\texttt{raghunathan@purdue.edu} \\
}

\begin{document}
\nocite{DBLP:journals/corr/Graves16}

\maketitle

\begin{abstract}
Deep Neural Networks (DNNs) have advanced the state-of-the-art in a variety of machine learning tasks and are deployed in increasing numbers of products and services. However, the computational requirements of training and evaluating large-scale DNNs are growing at a much faster pace than the capabilities of the underlying hardware platforms that they are executed upon. In this work, we propose  Dynamic Variable Effort Deep Neural Networks (DyVEDeep) to reduce the computational requirements of DNNs during inference. Previous efforts propose specialized hardware implementations for DNNs, statically prune the network, or compress the weights. Complementary to these approaches, DyVEDeep is a dynamic approach that exploits the heterogeneity in the inputs to DNNs to improve their compute efficiency with comparable classification accuracy. DyVEDeep equips DNNs with \emph{dynamic effort} mechanisms that, in the course of processing an input, identify how critical a group of computations are to classify the input. DyVEDeep dynamically focuses its compute effort only on the critical computations, while skipping or approximating the rest. We propose 3 effort knobs that operate at different levels of granularity \emph{viz.} neuron, feature and layer levels. We build DyVEDeep versions for 5 popular image recognition benchmarks --- one for CIFAR-10 and four for ImageNet (AlexNet, OverFeat and VGG-16, weight-compressed AlexNet). Across all benchmarks, DyVEDeep achieves 2.1$\times$-2.6$\times$ reduction in the number of scalar operations, which translates to 1.8$\times$-2.3$\times$ performance improvement over a Caffe-based implementation, with $<0.5\%$ loss in accuracy. 
\end{abstract}

\section{Introduction}

Deep Neural Networks (DNNs) have greatly advanced the state-of-the-art on a variety of machine learning tasks from different modalities including image, video, text, and natural language processing. However, from a computational standpoint, DNNs are highly compute and data intensive workloads. For example, DNN topologies that have won the ImageNet Large-Scale Visual Recognition Contest (ILSVRC) for the past 5 years, contain between 60-150 million parameters and require 2-20 giga operations of compute to classify a single image. These requirements are only projected to increase in the future, as data sets of larger sizes and topologies of larger complexity (more layers, features and feature sizes) are actively explored. Indeed, the growth in computational requirements of DNNs has far outpaced improvements in the capabilities of commodity computational platforms in recent years.

Two key scenarios exemplify the computational challenges imposed by DNNs: (i) Large-scale training, in which DNNs are trained on massive data-sets using high-performance server clusters or in the cloud, and (ii) Low-power inference, in which DNN models are evaluated on energy-constrained platforms such as mobile and deeply-embedded (Internet-of-Things) devices. Towards addressing the latter challenge, we propose Dynamic Variable Effort Deep neural networks (DyVEDeep), a new dynamic approach to improve the computational efficiency of DNN inference. 

{\bf \noindent Related Research Directions.} Prior research efforts to improve the computational efficiency of DNNs can be classified into 4 broad directions. The first comprises parallel implementations of DNNs on commercial multi-core and GPGPU platforms. Parallelization strategies such as model, data and hybrid parallelism (\cite{DBLP:journals/corr/Krizhevsky14,DBLP:journals/corr/0002AMVSKKD16}), techniques such as asynchronous SGD (\cite{40565}) and 1-bit SGD (\cite{1bit}) to alleviate communication overheads  are representative examples. The next set of efforts design specialized hardware accelerators to realize DNNs, trading off programmability, the cost of specialized hardware and design effort for efficiency. A spectrum of architectures ranging from low-power IP cores to large-scale systems have been proposed (\cite{5981829,Chen:2014:DMS:2742155.2742217,TPU}). The third set of efforts focus on developing new device technologies whose characteristics intrinsically match the computational primitives in neural networks, leading to improvements in energy efficiency (\cite{Liu:2015:RHR:2744769.2744900, Ramasubramanian:2014:SSD:2627369.2627625}). The final set of efforts exploit the fact that DNNs are typically over-parametrized (\cite{DBLP:conf/nips/DenilSDRF13}) due to the non-convex nature of the optimization space (\cite{DBLP:journals/corr/abs-1207-0580}). Therefore, they approximate DNNs by statically pruning network connections, representing weights with reduced bit precision and/or in a compressed format, thereby improving compute efficiency for a negligible loss in classification accuracy (\cite{DBLP:conf/nips/CunDS89,DBLP:journals/corr/HanPTD15,DBLP:conf/interspeech/LiuZW14,Venkataramani:2014:AEN:2627369.2627613,DBLP:conf/icassp/AnwarHS15,DBLP:conf/icassp/TanS16}).

{\bf \noindent DyVEDeep: Motivation and Concept.} In contrast to the above efforts, our proposal, {\bf Dy}namic {\bf V}ariable {\bf E}ffort {\bf Deep} neural networks (DyVEDeep~\footnote{The name stems from the notion that a network should "dive deep", or expend computational effort, judiciously as and where it is needed.}), leverages the heterogeneity in the characteristics of inputs to a DNN to improve its compute efficiency. The motivation behind DyVEDeep stems from the following key insights. 

First, in real-world data, not all inputs are created equal, {\em i.e.}, inputs vary considerably in their ``difficulty''. Intuitively, only  inputs that lie very close to the decision boundary require the full effort of the classifier, while the rest could be classified with a much simpler (\emph{e.g., linear}) decision boundary. In the context of DNNs, we can see that increasing network size provides a valuable, but nevertheless diminishing increase in accuracy. For example, in the context of ImageNet, increasing network's computational requirements by over 15$\times$ (from AlexNet to VGG) yields an additional 16\% increase in classification accuracy. This raises the question of whether some of the inputs can be classified with substantially fewer computations, while expending increased effort only for inputs that require it.

Second, for a given input, the effort needs to be expended across different parts of the network. For example, in an image recognition problem, the computations corresponding to neurons that operate on the image region where an object of interest is located are more critical to the classification output than the others. Also, some features may be less relevant than others in the context of a given input. For example, features that detect sharp edges may be less relevant if the current input is comprised mostly of curved surfaces.

Notwithstanding the above observations, state-of-the-art DNNs are static \emph{i.e.,} they are computationally agnostic to the nature of the input being processed and expend the same (worst case) computational effort on all inputs, which leads to significant inefficiency. DyVEDeep addresses this limitation by dynamically predicting which computations are critical to classify a given input and focusing compute effort only on those computations, while skipping or approximating the rest. In effect, the network expends computational effort on different subsets of computations for each input, reducing computational requirements in each case without sacrificing classification accuracy.

{\bf \noindent Dynamic Effort Knobs.} The key to the efficiency of DyVEDeep lies in favorably navigating the trade-off between the cost of identifying critical computations \emph{vs.} the benefits accrued by skipping or approximating computations. To this end, we identify three dynamic effort mechanisms at different levels of granularity \emph{viz.} neuron, feature and layer-levels. These mechanisms employ run-time criteria to dynamically evaluate the criticality of groups of computations and appropriately skip or approximate those that are deemed to be less critical.

\begin{itemize}
    \item {\bf Saturation Prediction and Early Termination (SPET)} operates at the neuron-level. It monitors the intermediate output of each neuron after processing a subset of its inputs (partial dot product between a subset of inputs and corresponding weights) and predicts the likelihood of the neuron eventually saturating after applying the activation function. If the partial sum is deep within the saturation regime (\emph{e.g.,} a large negative value in the case of ReLU), all further computations corresponding to the neuron are deemed to be non-critical and skipped.
    \item {\bf Significance-driven Selective Sampling (SDSS)} operates within each feature map, and exploits the spatial locality between neuron activations. A uniformly spatially sampled version of the feature is first computed. The activations of each remaining neuron is either approximated or accurately computed based on the magnitude and variance of its neighbors. 
    \item {\bf Similarity-based Feature Map Approximation (SFMA)} operates at the layer level, and examines the similarity between neuron activations in each feature map. If all neuron activations are similar, the convolution operation on the feature map is approximated by a single scalar multiplication of the average neuron activation value with the precomputed sum of kernel weights.
\end{itemize}

We develop a systematic methodology to identify the hyper-parameters for each of these mechanisms during the training phase for any given DNN. We built DyVEDeep versions for 5 popular DNN benchmarks \emph{viz.} CIFAR-10, AlexNet, OverFeat-accurate, VGG-16 and a weight-compressed AlexNet model. Our experiments demonstrate that by dynamically exploiting the heterogeneity across inputs, DyVEDeep achieves 2.1$\times$-2.6$\times$ reduction in the total number of scalar operations for $<$0.5\% loss in classification accuracy. The reduction in scalar operations translates to 1.8$\times$-2.3$\times$ improvement in performance in our software implementation of DyVEDeep using the Caffe deep learning framework on an Intel Xeon 2.7GHz server with 128GB memory.

The rest of the paper is organized as follows. Section~\ref{sec:relwork} describes prior research efforts related to DyVEDeep. Section~\ref{sec:dyvedeepApproach} details the proposed dynamic effort mechanisms and how they are integrated in DyVEDeep. Section~\ref{sec:exptmeth} outlines the methodology used in our experiments. The experimental results are presented in Section~\ref{sec:results}, and Section~\ref{sec:conclusion} concludes the paper. 

\section{Related Work} \label{sec:relwork}

In this section, we provide a brief summary of prior research efforts related to DyVEDeep, and highlight the distinguishing features of our work. Prior research on improving the computational efficiency of DNNs follows 4 distinct directions. 

The first class of efforts focus on parallelizing DNNs on commercial multi-cores and GPGPU platforms. Different work distribution strategies such as model, data and hybrid parallelism (\cite{DBLP:journals/corr/Krizhevsky14,DBLP:journals/corr/0002AMVSKKD16}), and hardware transparent on-chip memory allocation/management schemes such as virtualized DNNs (\cite{DBLP:journals/corr/RhuGCZK16}) are representative examples. The second class of efforts design specialized hardware accelerators that realize the key computation kernels in DNNs. A range of architectures targeting low-power mobile devices (\cite{5981829}) to high-performance server clusters (\cite{Chen:2014:DMS:2742155.2742217,TPU}) have been explored. The third set of efforts investigate new device technologies whose characteristics intrinsically match the compute primitives present in DNNs. Memristor-based crossbar array architectures (\cite{Liu:2015:RHR:2744769.2744900}) and spintronic neuron designs (\cite{Ramasubramanian:2014:SSD:2627369.2627625}) are representative examples.

The final set of efforts improve efficiency by approximating computations in the DNN. DyVEDeep falls under this category, as we propose to dynamically skip or approximate computations based on their criticality in the context of a given input. Therefore, we describe the approaches that fall under this category in more detail. To this end, we classify these approaches into static \emph{vs.} dynamic optimizations.

{\bf \noindent Static Techniques} Almost all efforts that approximate computations in DNNs are static in nature \emph{i.e.,} they apply the same approximation uniformly across all inputs. Static techniques primarily reduce the model size of DNNs by using mechanisms such as pruning connections (\cite{DBLP:conf/nips/CunDS89,DBLP:journals/corr/HanPTD15,DBLP:conf/interspeech/LiuZW14}), reducing the precision of computations (\cite{Venkataramani:2014:AEN:2627369.2627613,DBLP:conf/icassp/AnwarHS15}), and storing weights in a compressed format (\cite{DBLP:journals/corr/HanMD15}). For example, in the context of fully connected layers, HashNets (~\cite{DBLP:journals/corr/ChenWTWC15}) use a hash function to randomly group weights into bins, which share a common parameter value, thereby reducing the number of parameters needed to represent the network. Deep compression (\cite{DBLP:journals/corr/HanMD15}) attempts to prune connections in the network by adding a regularization term during training, and removing connections with weights below a certain threshold. 

In the context of convolution layers,~\cite{DBLP:journals/corr/DentonZBLF14,DBLP:conf/bmvc/JaderbergVZ14} exploit the linear structure of the network to find a suitable low rank approximation. On the other hand,~\cite{DBLP:conf/cvpr/LiuWFTP15} propose sparse convolutional DNNs, wherein almost $90\%$ of the parameters in the kernels are zeroed out by adding a weight sparsity term to the objective function. In contrast,~\cite{DBLP:journals/corr/MathieuHL13} demonstrate that performing convolution in the Fourier domain can yield substantial improvement in efficiency. Finally,~/cite{DBLP:journals/corr/FigurnovVK15} propose perforated CNNs, in which only a subset of the neurons in a feature are evaluated. The neurons to be evaluated for each feature are determined statically at training time. 

{\bf \noindent Dynamic Techniques.} Dynamic optimizations adapt the computations that are approximated based on the input currently being processed. Dynamic techniques are more powerful than statically optimised DNNs, as they can capture additional input-dependent opportunities for efficiency that static methods lack. Notwithstanding this, very little focus has been devoted to developing dynamic DNN approximation techniques. One of the first efforts in this direction (\cite{DBLP:journals/corr/abs-1305-2982}), utilizes stochastic neurons to gate regions within the DNN. Along similar lines,~\cite{DBLP:conf/nips/BaF13} propose Standout, where the dropout probability of each neuron is estimated using a binary belief network. The dropout mask is computed for the network in one shot, conditioned on the input to the network.~\cite{DBLP:journals/corr/BengioBPP15} extends a similar idea, wherein the dropout distribution of each layer is computed based on the output of the preceding layer. 

The dynamic effort mechanisms proposed in DyVEDeep are qualitatively different from the aforementioned efforts. Rather than stochastically dropping computations, effort knobs in DyVEDeep exploit properties such as the saturating nature of activation to directly predict the effect of approximation on the neuron output. Further, prior dynamic approaches have only be been applied to fully-connected networks trained on small datasets. Their applicability to large-scale DNNs remains unexplored. On the other hand, DyVEDeep is naturally applicable to both convolutional and fully connected layers, and we demonstrate substantial benefits on large-scale networks for ImageNet.

\section{DyVEDeep: Design Approach and Dynamic Effort Knobs} \label{sec:dyvedeepApproach}

\begin{figure}[htb]
\centering 
\includegraphics[width=0.5\textwidth]{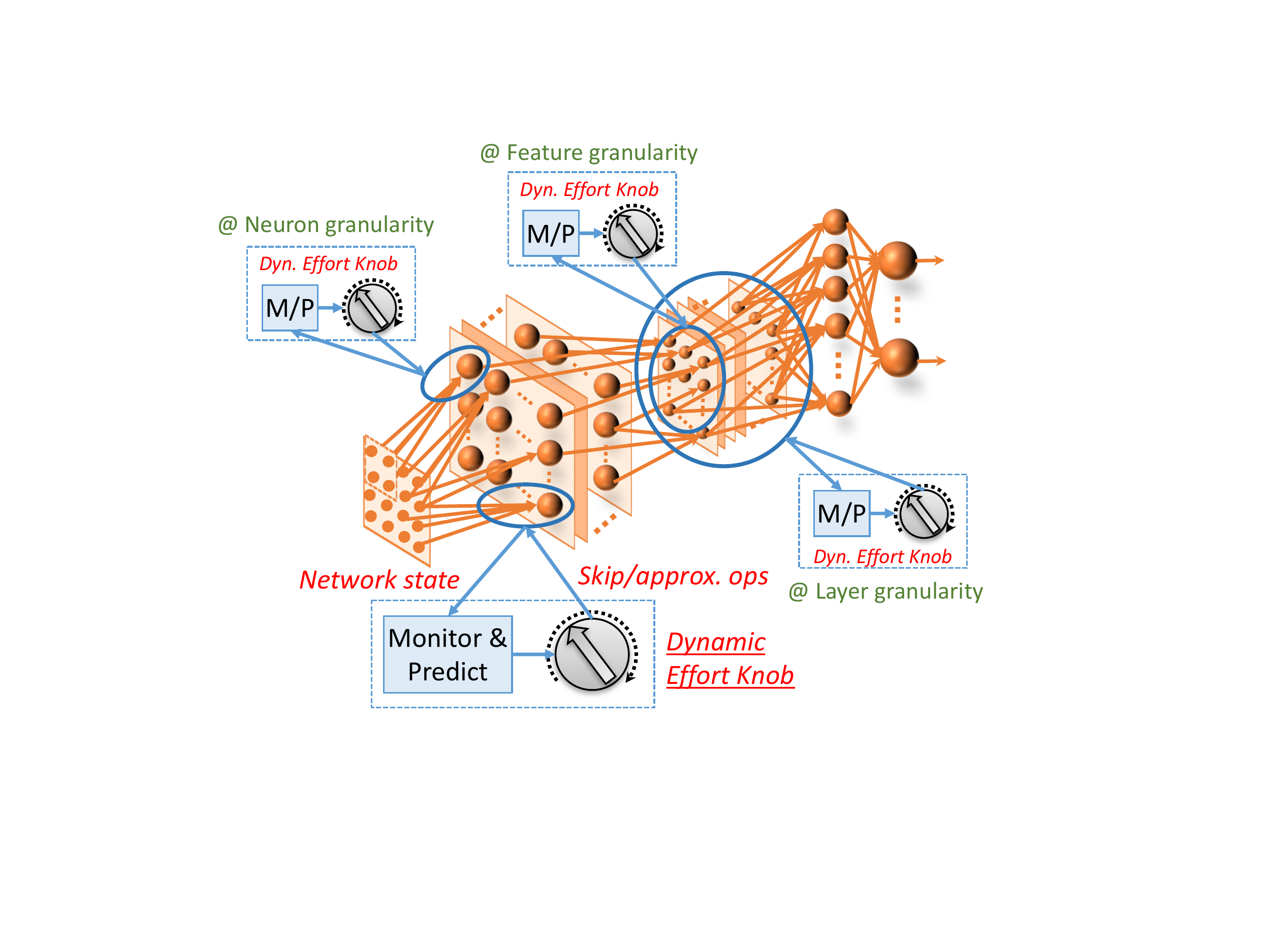}
\caption{DNN equipped with dynamic effort knobs}
\label{fig:dyvedeepTop}
\end{figure}

The key idea behind DyVEDeep is to improve the computational efficiency of DNNs by modulating the effort that they expend based on the input that is being processed. As shown in Figure~\ref{fig:dyvedeepTop}, we achieve this by equipping the DNN with dynamic effort mechanisms (``effort knobs") that dynamically predict criticality of groups of computations with very low overhead, and correspondingly skip or approximate them, thereby improving efficiency with negligible impact on classification accuracy. We identify three such dynamic effort mechanisms in DNNs that operate at different levels of granularity. We also propose a methodology to tune the hyper-parameters associated with these mechanisms so that variable effort versions of any DNN can be obtained with negligible loss in classification accuracy.

\subsection{Saturation Prediction and Early Termination}
Saturation Prediction and Early Termination (SPET) works at the finest level of granularity, which is at the level of each neuron in the DNN. In this case, we leverage the fact the almost all convolutional and fully connected layers are followed by an activation function that saturates on at least one side. For example, the commonly used Rectified Linear Unit (ReLU) activation function saturates at one end by truncating the negative inputs to zero, while passing the positive inputs as is. 

\begin{figure}[htb]
\begin{center}
\includegraphics[width=0.5\textwidth]{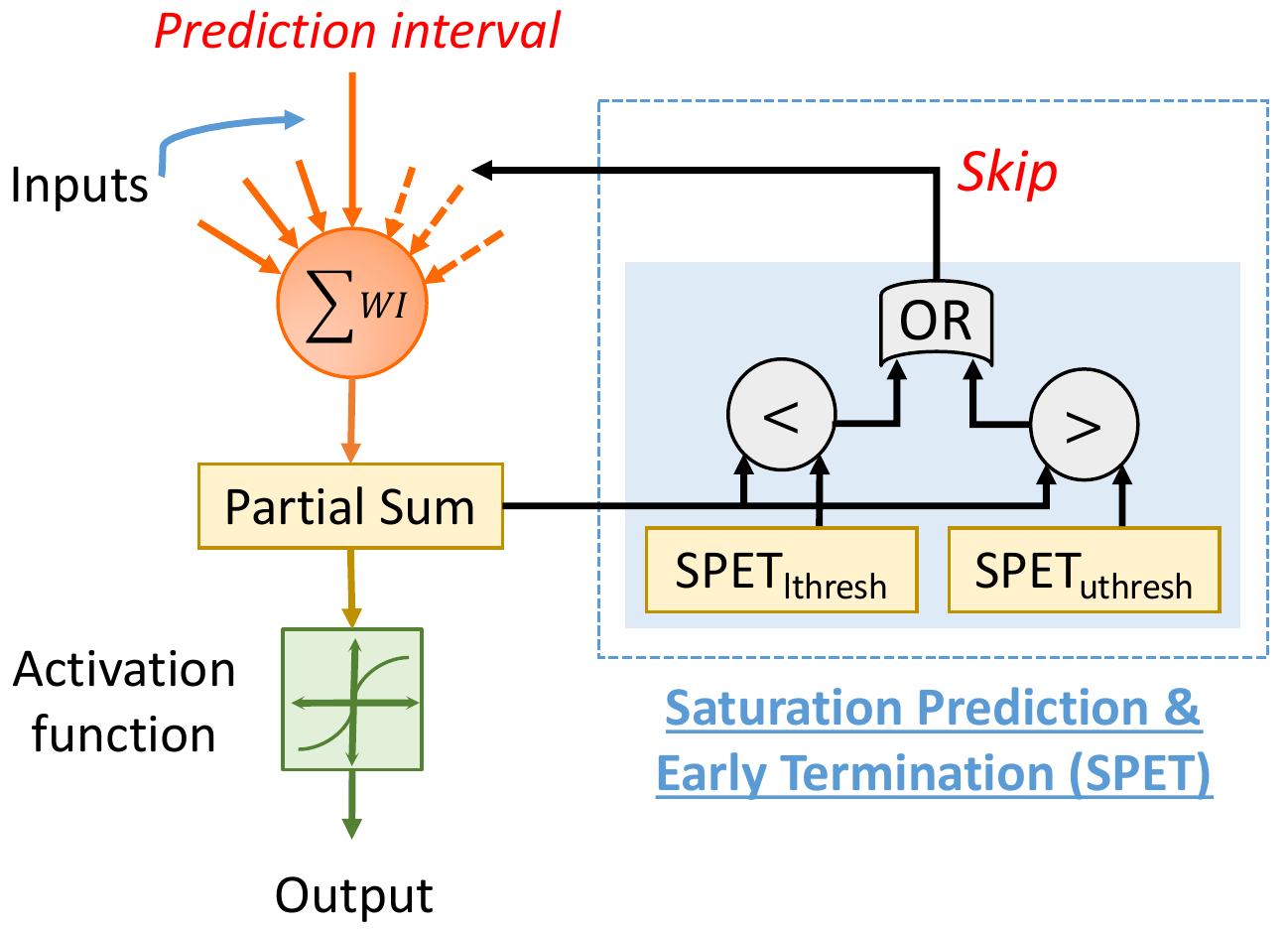}
\end{center}
\caption{Saturation Prediction and Early Termination}
\label{fig:neuknob}
\end{figure}

The key idea in SPET is that the \emph{actual value} of the weighted sum (dot product between a neuron's inputs and weights) does not impact the neuron's output, provided the sum will eventually cause the neuron's activation function to saturate. In the case of ReLU, it is unnecessary to compute the actual sum if it will eventually be a negative value, as any negative value would result in a neuron output of zero. Based on the above observation, as shown in Figure~\ref{fig:neuknob}, SPET monitors the partial weighted sum of a neuron after a predefined fraction of its inputs have been multiplied-and-accumulated. SPET then predicts whether the final partial sum would cause the neuron's activation function to saturate. To this end, we introduce the following hyper-parameters:
\begin{itemize}
    \item $SPET_{lThresh}$ and $SPET_{uThresh}$: We set two thresholds on the partial sum value of the each neuron. At the time of prediction, as shown in Equation~\ref{eqn:spetCond}, if the partial sum is found to be smaller than $SPET_{lThresh}$ or greater than $SPET_{uThresh}$, the partial sum computation is terminated early, and the appropriate saturated activation function value is returned as the neuron's output. If not, we continue to completely evaluate the partial sum value for the neuron.
\end{itemize}
\begin{equation}
    SPET_{out} = \left \{
    \begin{array} {l  l}
    $Terminate \& Saturate High$ & if $ Partial Sum $ > SPET_{uThresh} \\
    $Terminate \& Saturate Low$ & if $ Partial Sum $ < SPET_{lThresh} \\
    $Continue$ & otherwise
    \end{array} \right.
\label{eqn:spetCond}
\end{equation}
We note that if the activation function saturates in just one direction, only one of the SPET thresholds will be useful to predict saturation. For example, in the case of ReLU, only the $SPET_{lThresh}$ is used to predict saturation.

\begin{figure}[h]
\begin{center}
\includegraphics[width=14cm]{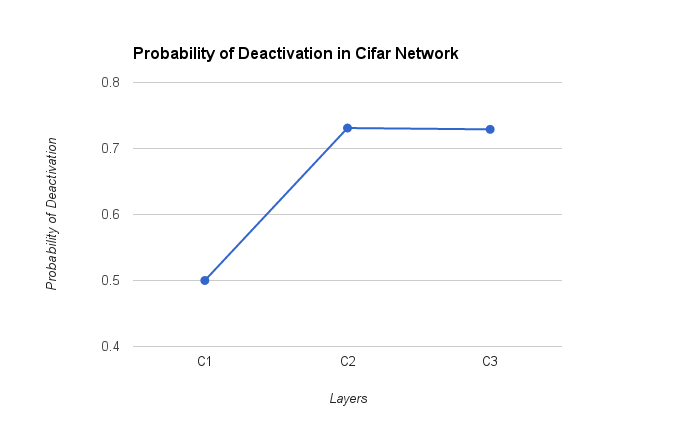}
\end{center}
\caption{Average fraction of neurons that saturate in each layer of the CIFAR-10 DNN}
\label{fig:DeactLayer}
\end{figure}

To demonstrate the potential benefits from SPET, Figure~\ref{fig:DeactLayer} shows the fraction of neurons in the convolutional layers of the CIFAR-10 DNN that saturate. We find that between 50\%-73\% of the neuron activations are zeros due to the ReLU activation function. Figure~\ref{fig:DeactLayer} also reveals that the fraction of neurons saturating increases as we proceed deeper into the network. We observed similar trends for larger networks such as AlexNet and OverFeat. Since a majority of neuron activations saturate in typical DNNs, SPET has a potential to achieve significant improvements in processing efficiency.

\begin{figure}[htb]
\begin{center}
\includegraphics[width=14cm]{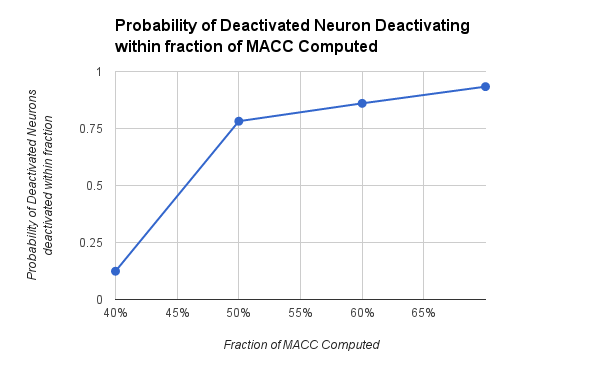}
\end{center}
\caption{Saturation prediction accuracy at different prediction intervals}
\label{fig:DeactSeries}
\end{figure}

{\bf \noindent Saturation Prediction Interval.} A key aspect of SPET is the interval at which we predict for saturation. On the one hand, predicting saturation after processing a small number of inputs to each neuron would frequently result in the prediction being incorrect, leading to a loss in classification accuracy. On the other hand, a larger prediction interval yields progressively smaller computational savings. Quantifying the above trade-off, Figure~\ref{fig:DeactSeries} illustrates, for the CIFAR-10 DNN, the fraction of neuron that were predicted to be saturated correctly at various prediction intervals. For the illustration in Figure~\ref{fig:DeactSeries}, we assume a $SPET_{lThresh}$ of 0 \emph{i.e.,} a neuron is predicted to saturate if its partial sum at the point of prediction is negative. We find that the fraction of neurons predicted correctly increases with the prediction interval.

The $SPET_{lThresh}$ and $SPET_{uThresh}$ hyper-parameters are determined during DNN training. We note that the prediction interval could also be learnt during the training process. However, we found that a simpler scheme where we fix the prediction interval at 50\% (\emph{i.e.,} we predict for saturation after half the inputs to a neuron have been processed) worked quite well in practice.

{\bf \noindent Rearranging Neuron Inputs.} For SPET to be most effective, the weights should be processed in decreasing order of magnitude, as larger weights are likely to have the most impact on the partial sum. However, this is not feasible in practise, as it affects the regularity in the memory access pattern, directly offsetting the savings from skipping computations. Also, in the case of convolutional layers, if the prediction interval is set to 50\%, inputs from half of the feature maps are ignored at the time of prediction. To maximize the range of inputs processed before prediction, while maintaining regularity in the memory access pattern, we rearrange the neuron inputs such that all odd indexed inputs are processed first, after which the prediction is made. The even indexed inputs are computed only if the neuron was not predicted to saturate. 

\subsection{Significance-driven Selective Sampling}

Significance-driven Selective Sampling (SDSS) operates the granularity of each feature in the convolutional layers of the DNN. SDSS leverages the spatial locality in neuron activations within each feature. For example, in the context of images, adjacent pixels in the input image frequently take similar values. As the neuron activations are computed by sliding the kernel over the image, the spatial locality naturally permeates to the feature outputs of convolutional layers. This behavior is also observed in deeper layers in the network. In fact, the saturating nature of the activation function enhances locality, as variations in the weighted sum between neighbors are masked if they both fall within the same saturation regime. 

\begin{figure}[htb]
\begin{center}
\includegraphics[width=0.5\textwidth]{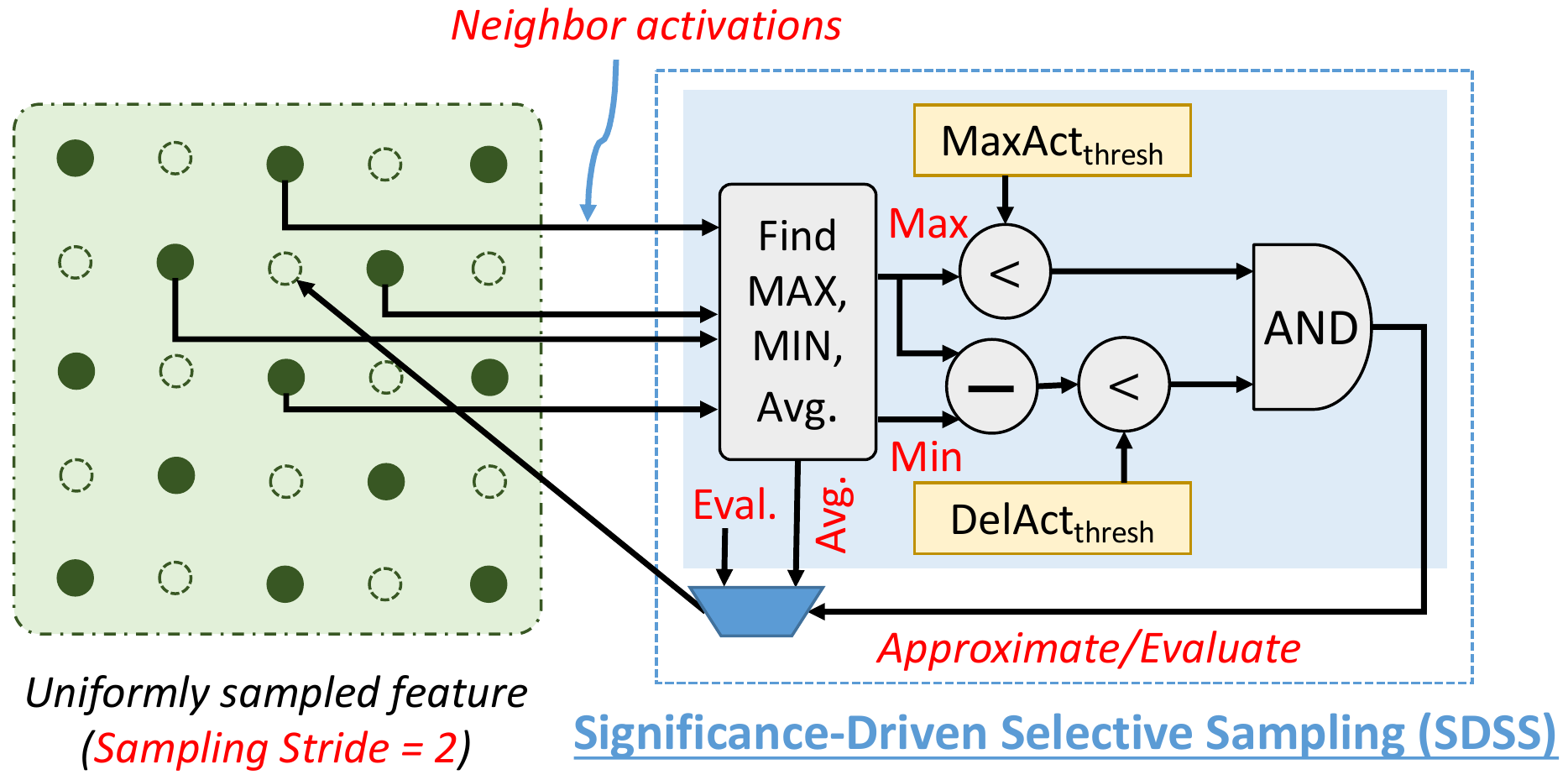}
\end{center}
\caption{Significance-driven Selective Evaluation}
\label{fig:feaknob}
\end{figure}

SDSS adopts a 2-step process to exploit the spatial locality within features. 

{\bf Uniform Feature Sampling.} In the first step, we  compute the activation values for a subset of neurons in the feature by uniformly sampling the feature. For this purpose, we define a parameter $SP$ that denotes the periodicity of sampling in each dimension. The value of $SP$ is chosen based on the size of the feature and the correlation between adjacent neuron activations. In our experiments, we used a sampling period of 2 across all convolutional layers in a DNN.

{\bf Significance-driven Selective Evaluation.} In the second step, as shown in Figure~\ref{fig:feaknob} we selectively approximate activation values of neurons that were not sampled in the first step. To this end, we define the following two hyper-parameters: (i) Maximum Activation Value Threshold ($MaxAct_{thresh}$), (ii) Delta Activation Value Threshold ($DelAct_{thresh}$). For each neuron in the feature that is yet to be computed, we examine the activation values of its immediate neighbors in all directions, and compute the maximum and range (difference between max and min) of the neighbors' activation values. If the maximum value is below the $MaxAct_{thresh}$ threshold and the range is less than the $DelAct_{thresh}$, then the activation value of the neuron is approximated to be the average of its neighbors. If not, the actual activation value of the neuron is evaluated.

Thus, the SDSS effort knob utilizes the magnitude and variance of neighbors to gauge whether a neuron lies within a region of interest, and accordingly expends computational effort to compute its activation value.

\subsection{Similarity-based Feature Map Approximation}

\begin{figure}[htb]
\begin{center}
\includegraphics[width=0.7\textwidth]{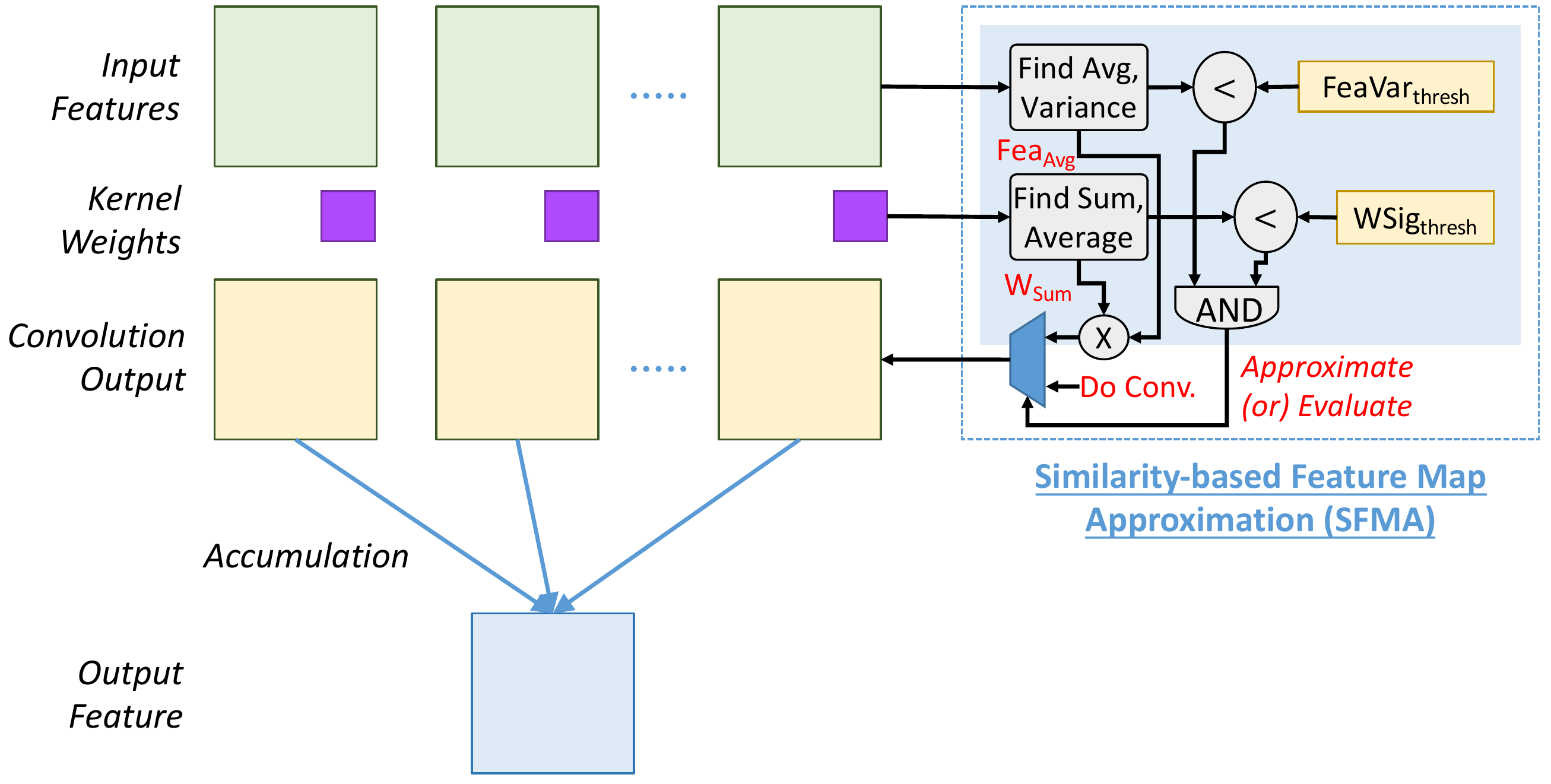}
\end{center}
\caption{Similarity-based Feature Map Approximation}
\label{fig:layerknob}
\end{figure}

Similarity-based Feature Map Approximation (SFMA) also exploits the correlation between activation values in a feature, but in a very different way. In SDSS, the spatial locality was exploited in computing the neuron activations themselves. In contrast, in the case of SFMA, the spatial locality is used to approximate computations that use the feature as their input. 
Consider a convolutional layer in which one of the input features has all of its neuron activations similar to each other. When a convolution operation is performed on this input feature by sliding the kernel matrix, all the entries in the convolution output are likely to be close to each other. Therefore, as shown in Figure~\ref{fig:layerknob}, we approximate entire convolution operation as follows. First, the average value of all neuron activations in the feature is computed. Next the sum of all weights in the kernel matrix is evaluated. We note that the sum can be precomputed and stored along with the kernel matrix. We then approximate all outputs of the convolution as the product of the average input activation and the sum of all kernel weights.

Mathematically, the above approximation can be expressed as follows. 
\[ConvOut_W =  \Sigma_{i=0}^{k^2} w_i * W_i = \mu * \Sigma_{i=0}^{k^2} w_i + \Sigma_{i=0}^{k^2} w_i * (W_i - \mu) \] 
\[ \approx \mu \Sigma_{i=0}^{k^2} w_{i} \]

In the above equation, $ConvOut_W$ is the convolution output for a window $W$ of size $k \times k$, where $k$ is the kernel size. $\mu$ is the mean of all the activation values in the feature. This approximation is valid when $\Sigma w_i * (W_i - \mu)$ is negligible.

To determine on which convolutions to apply the aforementioned approximation, we define the following 2 hyper-parameters:
\begin{itemize}
    \item Weight Significance Threshold ($WSig_{thresh}$) - We set this threshold on the sum of absolute values of the kernel weights. This is an approximate measure of significance of the current convolution to the output feature
    \item Feature Variance Threshold ($FeaVar_{thresh}$) - We set this threshold on the variance of the neuron activations in the feature. 
\end{itemize}
Given the hyper-parameters, the convolution is approximated when (i) the sum of the kernel weights are below $WSig_{thresh}$, indicating that the convolution is relatively less significant to the output feature, and (ii) the variance of neuron activations in the feature is below $FeaVar_{thresh}$, indicating that the error due to replacing the entire feature with its average is tolerable.

When the feature sizes are large, we do not check for the variance across the entire feature. Instead, we split the feature into multiple regions, that overlap on each dimension by the size of the kernel window. We check for variance within each region, and if the variance is below $FeaVar_{thresh}$, the kernel windows that fit entirely within the region are approximated.

\subsection{Integrating Effort Knobs}

We now describe how the different effort knobs---SPET, SDSS and SFMA---are combined in DyVEDeep. Since each effort knob operates at a different level of granularity, they can be easily integrated with each other. To combine SPET and SDSS, each neuron activation in the uniformly sampled features of SDSS are computed with SPET. However, we do not apply SPET to the neurons that are selectively computed in SDSS, as they are located in the midst of neurons with large activation values and/or variance, and are hence unlikely to saturate. SFMA fundamentally amounts to grouping a set of inputs (within a convolution window) to a neuron into a single input, and therefore directly fits with the process of evaluating a neuron with SPET/SDSS.

In summary, the SPET effort knob applies to both convolutional and fully connected layers of DNNs, and is most effective when majority of the neurons saturate. Since the convolutional layers towards the middle of the DNN have a large number of inputs per neuron and contain a substantial fraction of saturated neurons, we expect SPET to be most beneficial for those layers. The SDSS effort knob primarily applies only to convolutional layers, and is most effective when the features sizes are large. Therefore, the initial convolutional layers would benefit the most from SDSS. On the other hand, SFMA works best when there are a large number of features in the layer and when the feature sizes are small. Hence the middle and later convolutional layers are likely to benefit from SFMA.

\subsection{Hyper-parameter Tuning}

As described in the previous subsections, the dynamic effort knobs together contain 6 hyper parameters \emph{viz.} $SPET_{lThresh}$, $SPET_{uThresh}$, $MaxAct_{thresh}$, $DelAct_{thresh}$, $WSig_{thresh}$ and $FeaVar_{thresh}$. These hyper-parameters control how aggressively the effort knobs skip or approximate computations, thereby yielding a direct trade-off between computational savings \emph{vs.} classification accuracy. Using a pre-trained network and a training dataset, we systematically determine the DyVEDeep hyper-parameters before the DNN model is deployed. Ideally, we could define these parameters uniquely for each neuron in the DNN. For example, each neuron could have its unique $SPET_{lThresh}$ threshold to predict when it saturates (SPET), or $FeaVar_{thresh}$ threshold to deem if an input feature map can be approximated during its partial sum evaluation (SFMA). Clearly, this results in a prohibitively large hyper-parameter search space, and adds substantial overhead to the overall size of the DNN model. Since neurons in a given layer are computationally similar (same set of inputs, number of computations\emph{etc.}), we define the hyper-parameters at a layer-wise granularity \emph{i.e.,} all neurons within a layer share the same set of hyper-parameters. Also, since all our benchmarks utilized the ReLU activation function, we ignored the $SPET_{uThresh}$ when identifying the hyper-parameter configuration. 

\begin{algorithm}[htb]
\SetAlgoLined
\KwResult{DyVEDeep Network $N'$}
 Start with pre-trained network $N$\;
 \For {l in ConvolutionalLayers} {
   BinarySearch (($WSig_{thresh}$, $FeaVar_{thresh}$), l)\;
   BinarySearch ($MaxAct_{thresh}$, l)\;
   BinarySearch ($DelAct_{thresh}$, l)\;
   BinarySearch ($SPET_{lThresh}$, l)\;
 }
 \For {l in FullyConnectedLayers} {  
   BinarySearch ($SPET_{lThresh}$, l)\;
 }
\caption{Hyperparameter tuning algorithm}
\label{algo:hyperParamSearch}
\end{algorithm}

Algorithm~\ref{algo:hyperParamSearch} shows the pseudocode for the  hyper-parameter tuning process. Empirically, we observed that parameters corresponding to each effort knob can be independently tuned. Therefore, we adopt a strategy wherein we first identify a range of possible values for each hyper-parameter. Since computational savings monotonically increase or decrease with the value of each parameter, we perform a greedy binary search on its range. The range of each parameter can be identified as follows. The $SPET_{lThresh}$ and $MaxAct_{thresh}$ parameters vary over the entire range of values the partial sum of neurons can take in a layer. However, we typically observe that zero is a good lower bound for these parameters, as ReLU sets all negative values to 0. The upper bound is determined by evaluating the DNN on each input in the training dataset and recording the maximum partial sum value for each layer. The other parameters $DelAct_{thresh}$, $WSig_{thresh}$ and $FeaVar_{thresh}$ are naturally lowered bounded by 0 as they are thresholds on absolute magnitudes. Similar to $SPET_{lThresh}$ and $MaxAct_{thresh}$, the upper limit of the other parameters are also estimated by evaluating the DNN on the training set. 

Given a hyper-parameter and its range, the highest possible value for the parameter yields the maximum computation savings but adversely affects the classification accuracy. On the other extreme, the lowest value of the parameter does not impact the classification accuracy. However, it yields no computation savings and in fact adds a penalty for criticality prediction. Therefore, we perform a binary search on the range to identify the highest value of the parameter that yields negligible loss in classification accuracy ($<$0.5\% in our experiments). In the case of SFMA, we observed that the two hyper-parameters ($FeaVar_{thresh}$ and $WSig_{thresh}$) need to be searched together. Since the range of $FeaVar_{thresh}$ is more coarser than $WSig_{thresh}$, we loop over the values of $FeaVar_{thresh}$, and search for possible values of $WSig_{thresh}$ in each case.

In summary, by embedding dynamic effort knobs into DNNs, DyVEDeep seamlessly varies computational effort across inputs to achieve significant computational savings while maintaining classification accuracy. 

\section{Experimental Methodology} \label{sec:exptmeth}

In this section, we describe the methodology used in our experiments to evaluate DyVEDeep.

{\bf Benchmarks.} To evaluate DyVEDeep, we utilized pre-trained DNN models available publicly on the Caffe Model Zoo (\cite{caffemodelzoo})  benchmark repository. This reinforces DyVEDeep's ability to adapt to any given trained network. We used the following 5 DNN benchmarks in our experiments: CIFAR-10 Caffe network (\cite{cifar-10network}) for the CIFAR-10 dataset (\cite{Krizhevsky09learningmultiple}), and AlexNet (\cite{NIPS2012_4824}), Overfeat-accurate (\cite{Sermanet_overfeat:integrated}), VGG-16 (\cite{DBLP:journals/corr/SimonyanZ14a}), and compressed AlexNet (\cite{DBLP:journals/corr/HanMD15}) for the ImageNet ILSVRC 2012 data set (\cite{DBLP:conf/cvpr/DengDSLL009}). The inputs for the ImageNet dataset are generated by using a $224 \times 224$ center crop of the images in the test set. We randomly selected 5\% of the test inputs and used it as a validation set to tune the hyper parameters. We report speedup and classification accuracy results on the remaining 95\% of the test inputs. 


{\bf Performance Measurement.} We implemented DyVEDeep in C++ within the Caffe deep learning framework (\cite{jia2014caffe}). However, we could not directly integrate DyVEDeep within Caffe, as it composes all computations within a layer for a given batch size into a single GEMM (GEneral Matrix Multiplication) operation, which is offered by BLAS (Basic Linear Algebra Subprograms) libraries. BLAS libraries specifically optimize matrix operations at the assembly level. Since DyVEDeep requires more fine-grained computation skipping/approximation, we were unable to directly incorporate it within these routines. Therefore, we prototyped our own implementation for the convolutional layers within Caffe and used it in our experiments.

Our experiments were conducted on an Intel Xeon server operating at 2.7GHz frequency and 128GB memory. We added performance counters to both DyVEDeep and the baseline DNN implementation to measure the software execution time. All our timing results are reported for a single-threaded sequential execution. Also, for our experiments, we introduced dynamic effort knobs only in the convolutional layers of the DNN, as they dominated the overall runtime for all our benchmarks. However, we note that the reported execution times and performance benefits include the time taken by all layers in the network.

\section{Results} \label{sec:results}

In this section, we present the results of our experiments that demonstrate the benefits of DyVEDeep.

\subsection{Improvement in Scalar Operations and Execution Time}

\begin{figure}[htb]
\begin{center}
\includegraphics[width=0.5\textwidth]{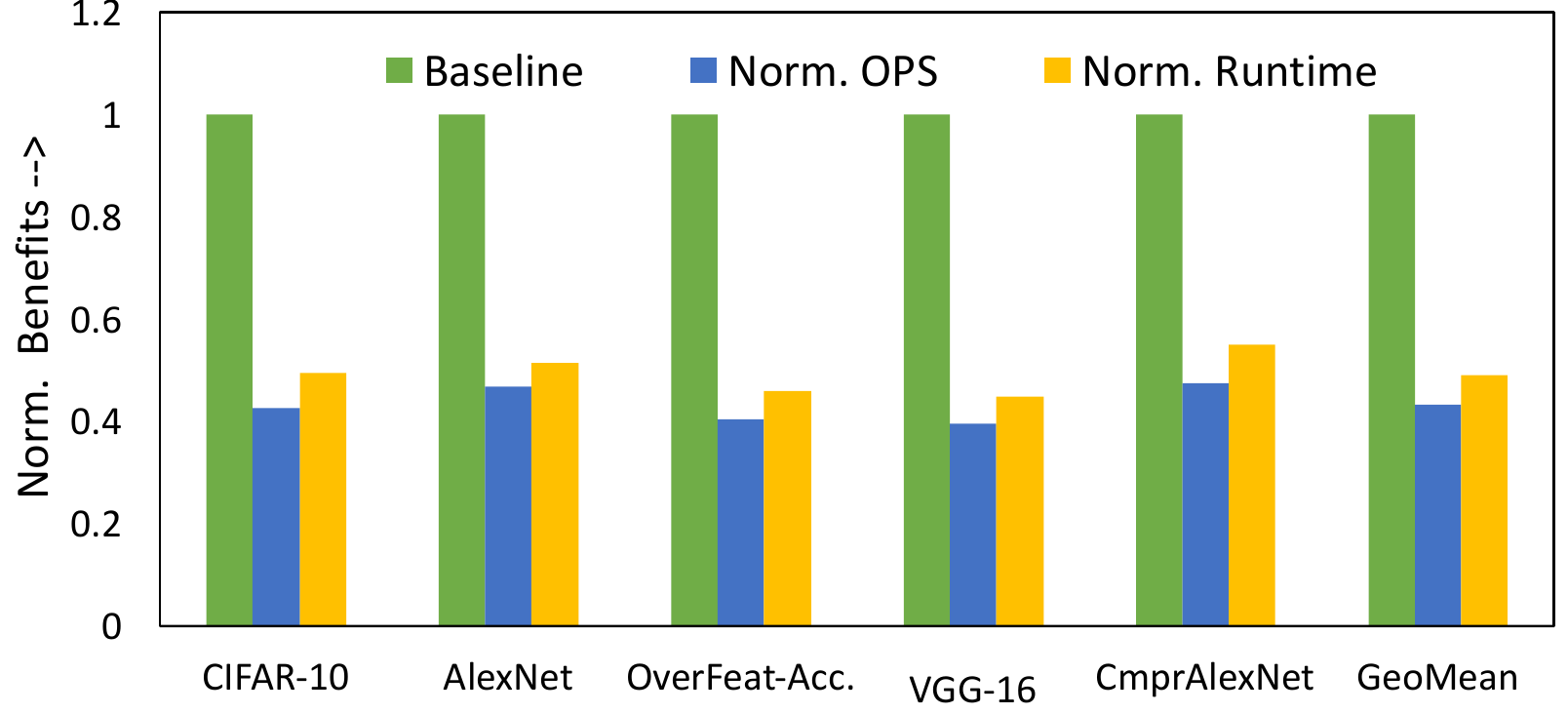}
\end{center}
\caption{Normalized improvement in scalar operations and execution time}
\label{fig:resNormBenefits}
\end{figure}

We first present the reduction in scalar operations and execution time achieved by DyVEDeep in Figure~\ref{fig:resNormBenefits}. Please note that the Y-axis in Figure~\ref{fig:resNormBenefits} is a normalized scale to represent the benefits in both scalar operations and runtime. We find that, across all benchmarks, DyVEDeep consistently achieves substantial reduction in operation count, ranging between 2.1$\times$-2.6$\times$. This translates to 1.8$\times$-2.3$\times$ benefits in software execution time. In all the above cases, the difference in classification accuracy between the baseline DNN and DyVEDeep was $<$0.5\%. On an average, the runtime overhead of the dynamic effort knobs in DyVEDeep was 5\% of the baseline DNN. Also, while the runtime benefits with DyVEDeep are quite significant, they are smaller compared to the reduction in scalar operations. This is expected as applying the knobs require us to alter memory access patterns and perform additional book keeping operations. Also, control operations, such as loop counters \emph{etc.}, that are inherent to any software implementation limits the fraction of runtime DyVEDeep can benefit. 

\subsection{Layer-wise and Knob-wise Breakdown of Compute Savings}

\begin{figure}[htb]
\begin{center}
\includegraphics[width=14cm]{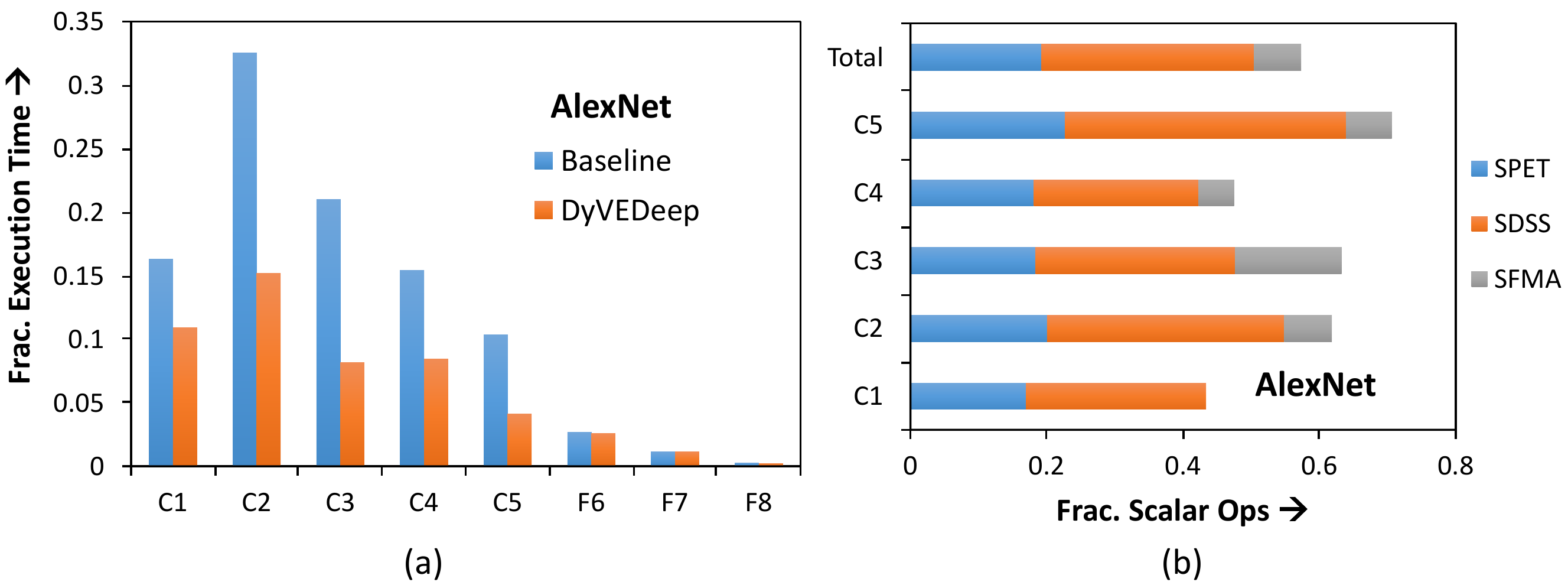}
\end{center}
\caption{(a) Layer wise breakdown of runtime benefits in AlexNet (b) Contribution of each effort knob to ops improvement}
\label{fig:AlexBreak}
\end{figure}

Figure~\ref{fig:AlexBreak}a shows the break down of run time savings across different layers of AlexNet, with the layers plotted on the X-axis and the average run time per layer normalized to the total baseline DNN run time on the Y-axis. We achieve 1.5$\times$ reduction in run time in the initial convolutional layers (C1,C2), which increases to 2.6$\times$ in the deeper convolutional layers (C3-C5). The C1 layer in AlexNet has a kernel size of 11$\times$11 and operates with a stride of 4. Hence, its output is less likely to have the correlation that SSDS expects. Also, since there are very few input features, SFMA is also not very effective. Also, the fraction of neurons saturating is relatively small in the first layers, which impacts the effectiveness of SPET. Hence, we achieve better savings in the deeper convolutional layers compared to the initial ones.

Figure~\ref{fig:AlexBreak}b compares the contribution of each effort knob to the overall savings for each convolutional layer in AlexNet. Over all layers, the SDSS knob yields the highest savings, reducing 31\% of the total scalar operations. The SPET and SFMA knobs contribute 19\% and 7\% respectively. We find that the effectiveness of each knob is more pronounced in the deeper convolutional layers.

\subsection{Visualisation of Effort Map of DyVEDeep}

\begin{figure}[htb]
\begin{center}
\includegraphics[width=3cm]{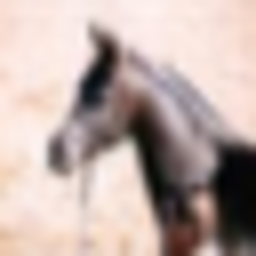}
\includegraphics[width=3cm]{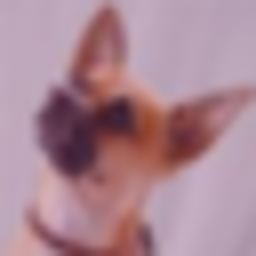}
\end{center}
\caption{Two sample images horse and dog from the CIFAR-10 data set to visualize the effort map of DyVEDeep}
\label{fig:horse-dog}
\end{figure}

\begin{figure}[htb]
\begin{center}
\includegraphics[width=6cm]{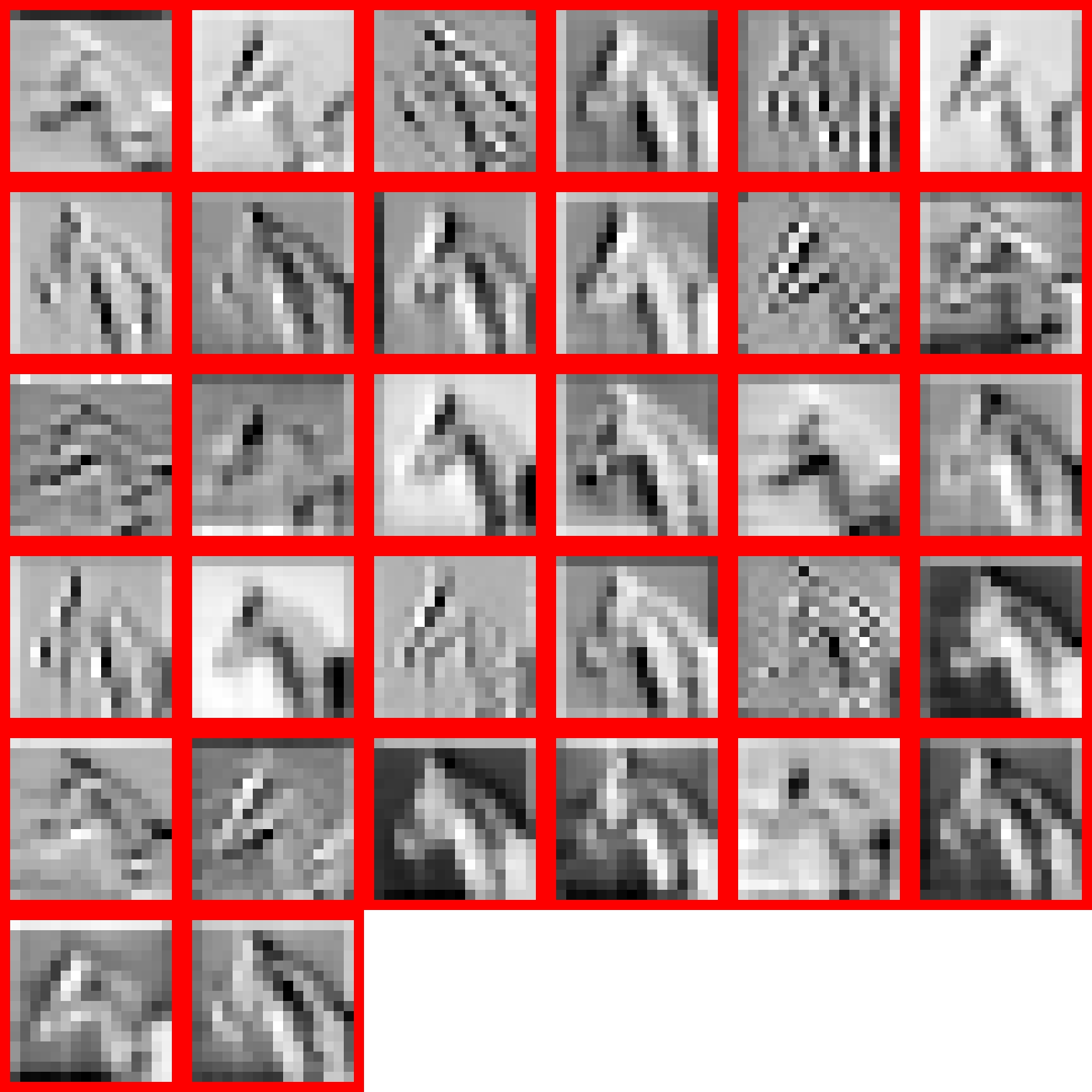} \includegraphics[width=6cm]{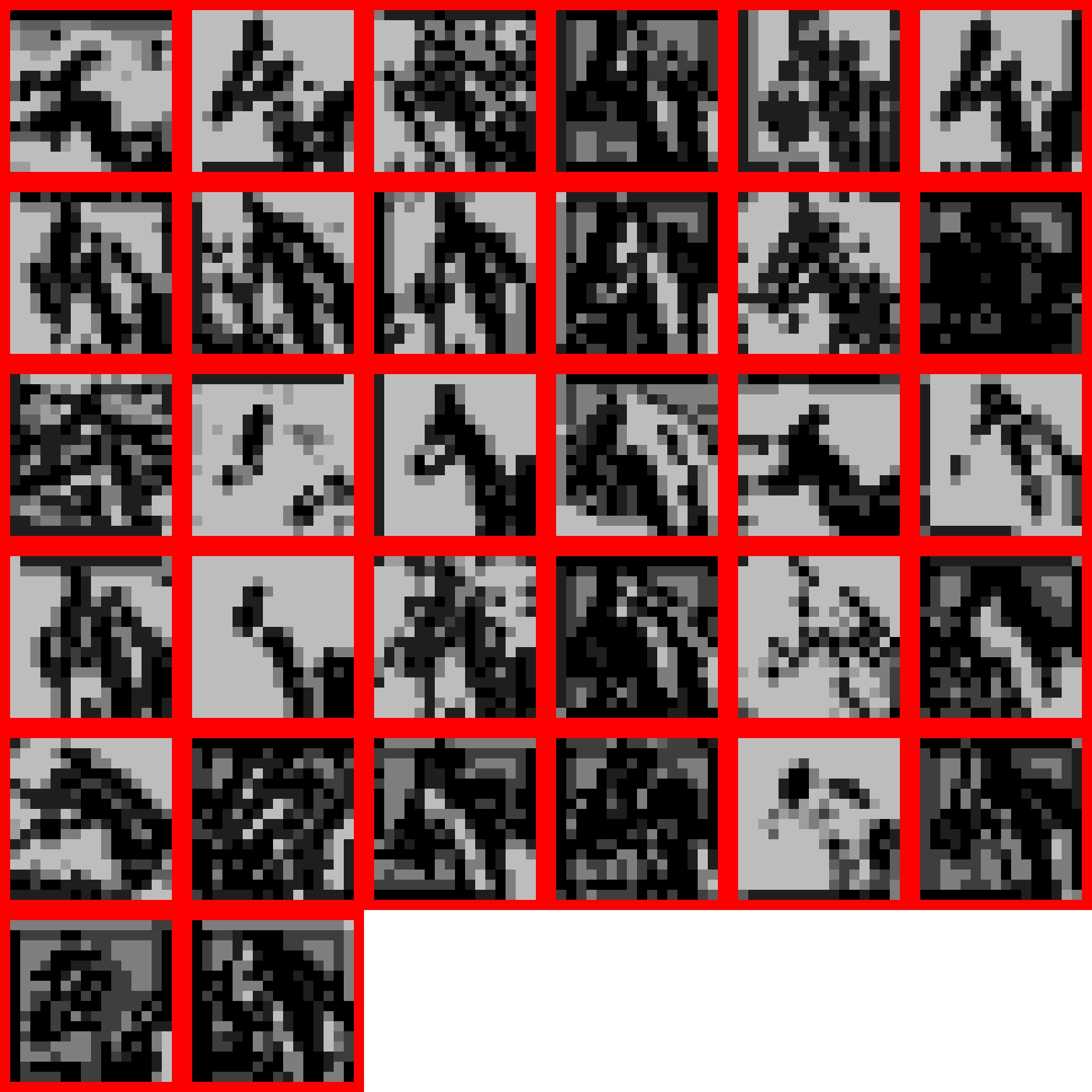}
\end{center}
\caption{Comparing the activation map and effort map of DyVEDeep for features in CIFAR-10 network layer C1 for the horse input}
\label{fig:horsemap}
\end{figure}

\begin{figure}[htb]
\begin{center}
\includegraphics[width=6cm]{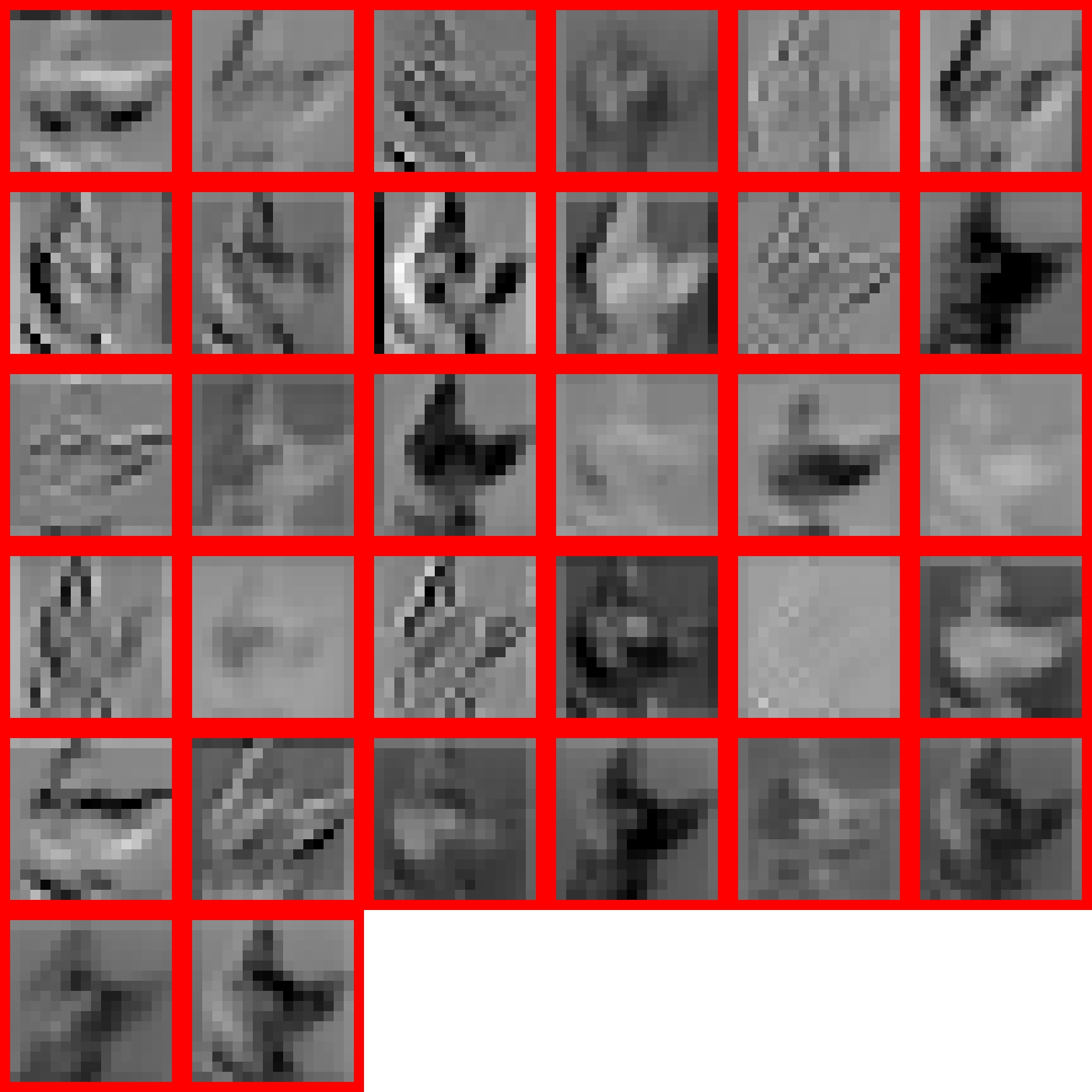} \includegraphics[width=6cm]{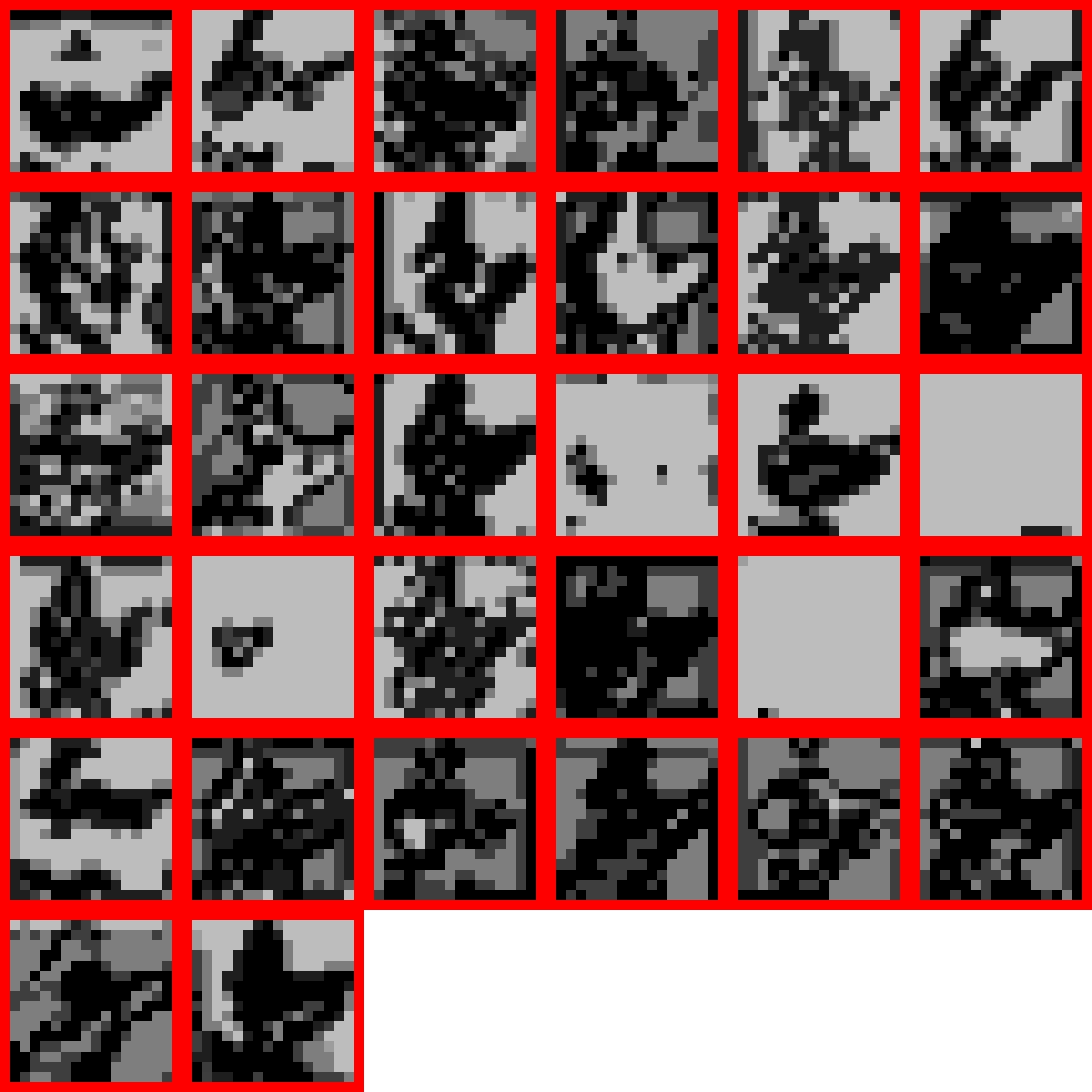}
\end{center}
\caption{Comparing the activation map and effort map of DyVEDeep for features in CIFAR-10 network layer C1 for the dog input}
\label{fig:catmap}
\end{figure}

Figures~\ref{fig:horsemap} and~\ref{fig:catmap} illustrate the normalised effort map of DyVEDeep for all features in layer C1 for two sample images (Figure~\ref{fig:horse-dog}) from the CIFAR-10 data set. We use layer C1, as this is the closest layer to the actual image and allows for better visualization. The normalization is done with respect to the number of operations that would have been performed to compute the neuron, had our knobs not been in place. Darker regions represent more computations. It is remarkable to see that DyVEDeep focuses more effort on precisely the regions of the image, that contains the object of interest. We compare this with the activation map of the corresponding features. Here, the darker regions represent activated neurons. This has been done to highlight the correlation between the activation values and the effort that DyVEDeep expends on the corresponding neurons. 
The activation map demonstrates that regions where the activation value of neurons are high have a higher variance in the values, that makes it harder to approximate them. However, the $DelAct_{thresh}$ parameter ensures that DyVEDeep constrains the effort spent in regions with uniform activation values. These effort maps corroborate our knobs' effectiveness in identifying the critical computations for the current input.

\section{Conclusion}  \label{sec:conclusion}
Deep Neural Networks have significantly impacted the field of machine learning, by enabling state-of-the-art functional accuracies on a variety of machine learning problems involving image, video, text, speech and other modalities. However, their large-scale structure renders them compute and data intensive, which remains a key challenge. We observe that state-of-the-art DNNs are static \emph{i.e.} they perform the same set of computations on all inputs. However, in many real-world datasets, there exists significant heterogeneity in the compute effort required to classify each input. Leveraging this opportunity, we propose Dynamic Variable Effort Deep Neural Networks (DyVEDeep), or DNNs that modulate their compute effort dynamically ascertaining which computations are critical to classify a given input. We build DyVEDeep versions of 4 popular image recognition benchmarks. Our experiments demonstrate that DyVEDeep achieves 2.1$\times$-2.6$\times$ reduction in scalar operations and 1.9$\times$-2.3$\times$ reduction in runtime on a Caffe-based sequential software implementation, while maintaining the same level of classification accuracy.

\bibliographystyle{iclr2017_conference}
\bibliography{iclr2017_conference}

\end{document}